\documentclass{article}

\usepackage[nonatbib, final]{neurips_2020}

\usepackage[utf8]{inputenc} 
\usepackage[T1]{fontenc}    
\usepackage{hyperref}       
\usepackage{url}            
\usepackage{booktabs}       
\usepackage{multirow}
\usepackage{array}
\usepackage{amsfonts}       
\usepackage{nicefrac}       
\usepackage{microtype}      
\usepackage{graphicx}
\usepackage{caption}
\usepackage{subcaption}
\usepackage{enumitem}
\usepackage{appendix}
\usepackage{placeins}
\usepackage{flafter}

\hypersetup{colorlinks=true, citecolor=green, linkcolor=red}

\newcommand{\etal}{\textit{et al}.}
\newcommand{\ie}{\textit{i}.\textit{e}.}

\makeatletter
\newcommand{\footnoteref}[1]{%
\ltx@ifpackageloaded{hyperref}{
  \ifHy@hyperfootnotes
    \hbox{\hyperref[#1]{%
            \@textsuperscript {\normalfont \ref*{#1}}}}%
  \else
    \hbox{\@textsuperscript {\normalfont \ref*{#1}}}%
  \fi%
}{
    \hbox{\@textsuperscript {\normalfont \ref{#1}}}%
 }%
}
\makeatother

\title{Reinforcement Learning Agents for Ubisoft's Roller Champions}

\author{%
  Nancy Iskander\thanks{Work done while at Ubisoft.} \\
  Riot Games\\
  \texttt{niskander@riotgames.com} \\
  \And
  Aurelien Simoni \\
  Ubisoft \\
  \texttt{aurelien.simoni@ubisoft.com} \\
  \AND
  Eloi Alonso \\
  Ubisoft \\
  \texttt{eloi.alonso@ubisoft.com} \\
  \And
  Maxim Peter \\
  Ubisoft \\
  \texttt{maxim.peter@ubisoft.com} \\
}

\begin{document}

\maketitle

\begin{abstract}
  In recent years, Reinforcement Learning (RL) has seen increasing popularity in research and popular culture. However, skepticism still surrounds the practicality of RL in modern video game development. In this paper, we demonstrate by example that RL can be a great tool for Artificial Intelligence (AI) design in modern, non-trivial video games. We present our RL system for Ubisoft's Roller Champions, a 3v3 Competitive Multiplayer Sports Game played on an oval-shaped skating arena. Our system is designed to keep up with agile, fast-paced development, taking 1--4 days to train a new model following gameplay changes. The AIs are adapted for various game modes, including a 2v2 mode, a Training with Bots mode, in addition to the Classic game mode where they replace players who have disconnected. We observe that the AIs develop sophisticated co-ordinated strategies, and can aid in balancing the game as an added bonus. Please see the accompanying video at \href{https://vimeo.com/466780171}{https://vimeo.com/466780171}~\footnote{\label{pass}Password: rollerRWRL2020} for examples.

\end{abstract}


\section{Introduction and Related Work}

Reinforcement Learning (RL) is an active topic of research that has received a lot of attention in the literature~\cite{Vinyals2019}~\cite{schulman2017proximal}~\cite{baker2019emergent}~\cite{tomasev2020assessing}. However, the role of RL in AI design is still unclear in the video games industry, and is highly doubted by some~\cite{rlblogpost}. By presenting our RL system for Roller Champions, we demonstrate that RL can be a powerful and practical AI design tool for modern, non-trivial video games. We tackle the challenges of agile development, real-time inference, competitive and collaborative gameplay, and multi-purpose AI where winrate is secondary to player experience.

Baker~\etal~\cite{baker2019emergent} observe emergent complexity in their Hide And Seek environment, and propose that multi-agent co-adaptation is capable of producing complex and intelligent behaviour. We observe analogous kinds of multi-agent co-ordinated co-dependent behaviours in Roller Champions (section ~\ref{sec:emergentbhvrs}).

The role of RL in balancing games is an open area of research. Recently, RL has been used to assess game balance in chess variants~\cite{tomasev2020assessing}. While aiding in game balance was not one of our key goals, we found that the AIs readily adapted to changes in game mechanics, and inadvertently revealed exploits (section~\ref{sec:balance}).

\section{Roller Champions}
At the time of writing, Roller Champions is set to be released by Ubisoft in early 2021. The Roller Champions official website~\cite{rollerWebsite} provides additional details.

\subsection{Gameplay}
Roller Champions: Rules of the Arena~\cite{rulesOfArena} describes the high-level gameplay. 

\renewcommand{\arraystretch}{1.3}

\subsection{Challenging application for RL}
\label{sec:challenges}
Roller Champions presents a number of characteristics that make it an interesting case-study as an application of RL in modern video game development, namely:

\paragraph{Collaborative.} Roller Champions is a collaborative game, where winning hinges upon the players' ability to pass the ball to each other, goal tend, and peel for the ball bearer. Most importantly, they need to coordinate and assume self-assigned roles depending on the situation.
\paragraph{Competitive.} The competitive gameplay further increases the complexity of Roller Champions. For instance, one mistake can cost the whole team all of its completed laps and checkpoints. Players need to weigh the advantage of completing additional laps for extra points against the risk of losing their progress due to the enemy team getting ahold of the ball.
\paragraph{Reward Sparsity.} In Roller Champions, players must pass a series of ordered checkpoints, while holding the ball, in order to activate the goal for their team, at which point only that team is allowed to score. The long sequence of events needed to score makes for sparse rewards.
\paragraph{Multi-purpose AI.} Our goal was to develop AI that is not just good at winning, but that would also be fun to play with and against. For \verb+Training With Bots+ mode, we wanted an all-around well-performing competitive AI that would still be fun. But in order to be able to use the AI to replace players who disconnect, we wanted another model that would be more fun than it is competitive. 
\paragraph{Live Game.} Roller Champions is set to be a free-to-play, live game. This means that the training environment has to be efficient enough to keep up with continuous gameplay and balance changes. 
\paragraph{Real-time Inference.} Since the AI is to be deployed in real games alongside human players, inference needs to be fast enough to maintain the target framerate. 


\section{Training Environment}
\label{sec:env}

Roller Champions is built in Unity, and the Unity ML-Agents Toolkit~\cite{unity} is used for training. Proximal Policy Optimization~\cite{schulman2017proximal} is used as the training algorithm. The policy network was kept relatively small for training and inference speed, containing 3 layers of 512 neurons (excluding input and output layers). We use a decision interval of 15 (\ie~observations are collected and an action is decided every 15 \verb+FixedUpdates+).

Self-play~\cite{bansal2018emergent} is used to obtain AIs that can play against a wide-range of strategies and skill levels. In the absence of penalties (see section~\ref{sec:rewards}), we found self-play useful in preventing collaboration between the two enemy teams.

\subsection{Observations}
\label{sec:observations}
Observations are derived directly from the game state. We experimented with two approaches:
\newcommand\litem[1]{\item{\bfseries #1.\enspace}}

\paragraph{Raycasts} Rays are cast from each agent's position at different angles to detect nearby objects and their types. Relevant game state observations are added.
\paragraph{Game Entities} Observations of relevant game entities are always included. These entities include: ball, enemies, allies, goal, checkpoints, and laps. For each entity, relative position, speed, and clear line-of-sight flag are observed. For players, flags indicating whether they are hurt, in air, or performing certain actions are added. The agent also observes its own position, state, and speed.

Even though the \verb+Raycasts+ approach gave satisfactory results, we eventually chose the \verb+Game Entities+ approach. Collecting game entity observations is much faster due to requiring a small number of raycasts (needed for line-of-sight flags only). The performance advantage is useful during both inference and training. 

In addition, the total number of observations can be kept low --at 60 and 78 total observations in \verb+2v2+ and \verb+3v3+ modes respectively-- without omitting any relevant information. A smaller observation size makes training easier and reduces the number of training steps required to learn a satisfactory policy. 

Observations of time and score are intentionally omitted. This makes it possible to reuse models in multiple game modes with different end conditions. The system is easily adapted to \verb+2v2+ mode by simply omitting two player observations.

\subsection{Action Space}
\label{sec:actionspace}
The action space is discrete and contains two branches: \verb+Agent Action+ and \verb+Agent Direction+.

\paragraph{Agent Action} Near one-to-one correspondence with actions available to human players: \verb+Accelerate, Pump, Brake, Throw, Pass, Call For Pass, Dash, Dodge, Jump+. \verb+Throw+ distance is fixed for AIs. The decision interval is chosen to be slightly longer than certain player actions in order to make it possible for the AIs to perform combos:\verb+Dive+: \verb+Dash+$\,\to\,$\verb+Dash+, \verb+Dodge Dive+: \verb+Dodge+$\,\to\,$\verb+Dash+, \verb+Uppercut+: \verb+Dash+$\,\to\,$\verb+Jump+. 

\paragraph{Agent Direction} Controls agent's rotation \emph{only}, except with \verb+Throw+, where it is used as throw direction. Initially, standard 2D cartesian coordinates were used. This allowed the agents to navigate the arena well enough, but the amount of time it was taking them to learn the game was impractical. Therefore, we made the switch to directions that took the shape of the arena into account, namely: \verb+Clockwise+, \verb+Anti-Clockwise+, \verb+Up Inner Wall+, and \verb+Up Outer Wall+. These were further augmented with the following high-level directions: \verb+Ally1+, \verb+Ally2+, \verb+Enemy1+, \verb+Enemy2+, \verb+Enemy3+, \verb+Ball+, and \verb+Goal+. Compared to cartesian co-ordinates, this action space significantly reduced the time required for agents to learn the basics of the game.

\paragraph{Action Masking} Actions that cannot be performed are masked. \verb+Agent Actions+ can be masked due to cooldown, ball state, game state, etc.\ while \verb+Agent Directions+ are masked for targets that the observing agent does not have a clear line-of-sight to. Action masking was imperative to overcoming superstition~\cite{bontrager2019superstition} and for successful training. Before implementing action masking, training was unstable, eventually leading to a policy where the AIs did not perform any actions if penalties were used. 

\subsection{Rewards}
\label{sec:rewards}
Rewards are only given for progress towards winning a match, namely: Progressing a checkpoint (0.05 scaled by checkpoint index) and scoring to win the match (2 per point scored). 

\paragraph{Teamwork} Teamwork is encouraged through the whole team receiving the same reward or penalty. 
\paragraph{Penalties} Penalties are given to team A whenever team B receives a reward. The magnitude of the penalty has a huge effect on the resulting behaviours, and is controlled using a penalty multiplier parameter. 

Since agents can receive rewards and penalties due to actions of other players that do not directly affect them, they are susceptible to superstition~\cite{bontrager2019superstition}. Batch size is increased to prevent superstition. 

\subsection*{Tailoring rewards for custom behaviours}

\begin{figure}[!h]
  \centering
  \includegraphics[width=0.9\linewidth]{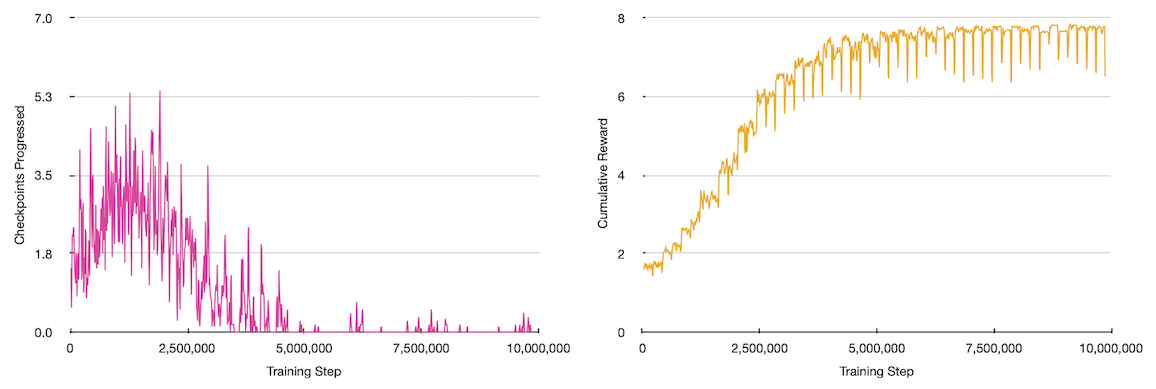}
  
  \caption{Assigning rewards for wall skating interferes with training. The reward was scaled by the agent's position on the wall, with a maximum value of 0.003 for the highest position, and was only given to the ball bearer, while grounded. Training is technically successful and agents are able to maximize their rewards, but they stop playing the game, opting to maximize the more immediate wall skate reward instead.}
  \label{fig:yreward}
\end{figure}

In a nutshell, tailoring rewards to obtain behaviours that are not directly related to win-rate did not work. There were two custom behaviours that we tried to encourage through rewards: Co-operation through the \verb+Pass+ action, and wall skating. In both cases, the rewards were exploited; the AIs quickly unlearned scoring, opting to maximize the more immediate rewards instead (see figure~\ref{fig:yreward}).

However, it is worthwhile to mention that the desired behaviours (\ie~wall skating and co-ordinated passes) automatically and consistently emerge using our current system (section~\ref{sec:emergentbhvrs}).

\subsection{Adapting the game for training}
\label{sec:optimize}

The game was adapted to run in server-only mode, without graphics or networking. Time scaling is used to speed up the simulation (accuracy of the simulation is not compromised due to game logic being implemented in Unity's \verb+FixedUpdate+). We typically run 3--15 simultaneous environments. Rather than using an entire match as an episode, shorter one-goal episodes are used. Tensorflow's GPU acceleration is used to speed up training.

With these adaptations, episodes initially conclude in about 30 seconds, reduced to about 15 seconds after the AIs learn how to score. 



\section{Results and Discussion}

\subsection{Performance}

\subsubsection*{Training Performance}

It takes roughly 10 hours per 10 million steps of training. The total number of training steps depends on the desired result; the AIs can improve their strategy (compared to previous policies) for up to 100 million steps, before hitting a plateau. At 12 million steps, the AIs have a high degree of tendency to clump together in one spot of the arena. At 20 million, they start to become more goal-oriented, utilizing the entire arena and completing laps faster and more frequently. At 30 million, co-ordinated co-dependent emergent behaviours are observed (section~\ref{sec:emergentbhvrs}). At 60 million+ steps, a distinct strategy emerges, with ELO continuing to rise at a slower rate.

In practice, the model obtained at ~30--40 million steps can be preferable to that obtained at 60 million steps due to the latter model being lower in entropy and thus more predictable. Overall, it only takes 1--4 days to produce a new model following gameplay or balance changes. This makes the system very practical for fast-paced, agile development, which is suitable for a live game. 

\subsubsection*{Inference Performance}

Framerate is throttled to 30 frames per second (FPS) in Roller Champions. This framerate is easily achieved with all players replaced by AIs, and can easily go up to 60 FPS without throttling. Inference efficiency can be attributed to the usage of a  small policy network and CPU-efficient observations derived directly from the game state (see section~\ref{sec:observations}). 

\subsection{Assessing Game Balance}
\label{sec:balance}

\begin{figure}[!h]
  \centering
  \includegraphics[width=0.8\linewidth]{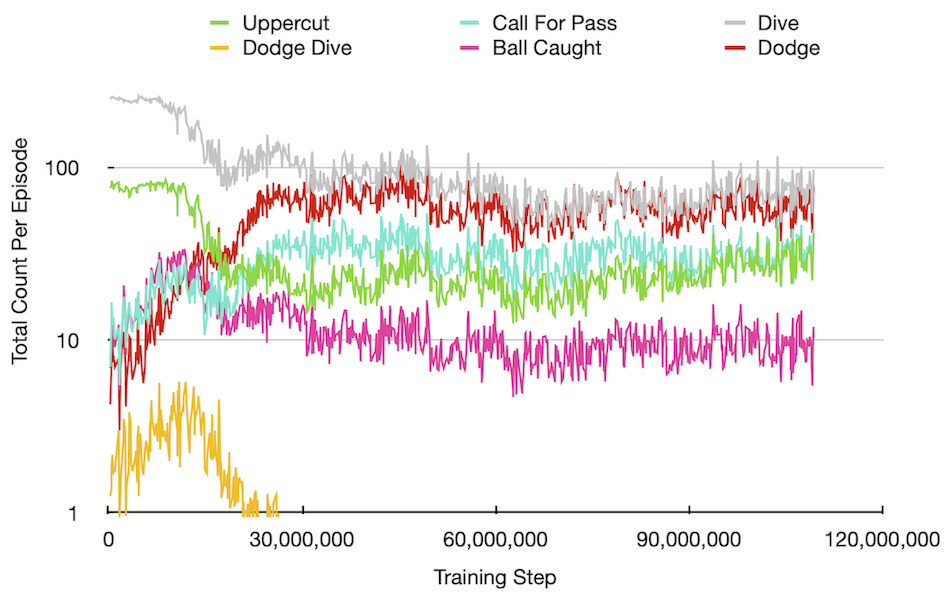}
  \caption{Player Action graphs demonstrate how the AIs evolve over the training period, and are helpful in assessing game balance. This graph indicates that the Dodge Dive combo is underpowered compared to the other two combos (Uppercut and Dive) and coordinated actions such as Passing. We speculate that the Tumble penalty caused by these combos is perhaps more penalizing when the player is holding the ball (as is the case for Dodge Dive).}
  \label{fig:playeractions}
\end{figure}

Observing how the AIs evolve over the training period can be helpful in assessing game balance. 
For instance, figure~\ref{fig:playeractions} shows that even though the \verb+Dodge+ as well as combos such as \verb+Uppercut+ and \verb+Dive+ are well utilized by the agents, the \verb+Dodge Dive+ combo is almost never used after the 30 million step mark, indicating that this action is underpowered compared to other actions.

\begin{figure}[!h]
     \centering
     \begin{subfigure}[b]{0.4\textwidth}
         \centering
         \includegraphics[width=\textwidth]{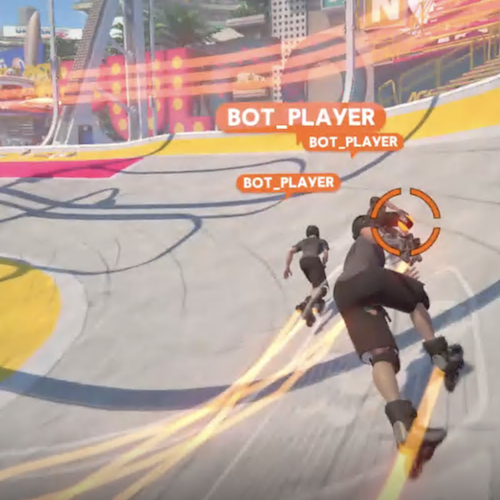}
         \caption{The Draft mechanic gave players bonus speed for skating in the wake of other players. This was exploited by the agents who always skated together.}
         \label{fig:3together}
     \end{subfigure}
     \hfill
     \begin{subfigure}[b]{0.4\textwidth}
         \centering
         \includegraphics[width=\textwidth]{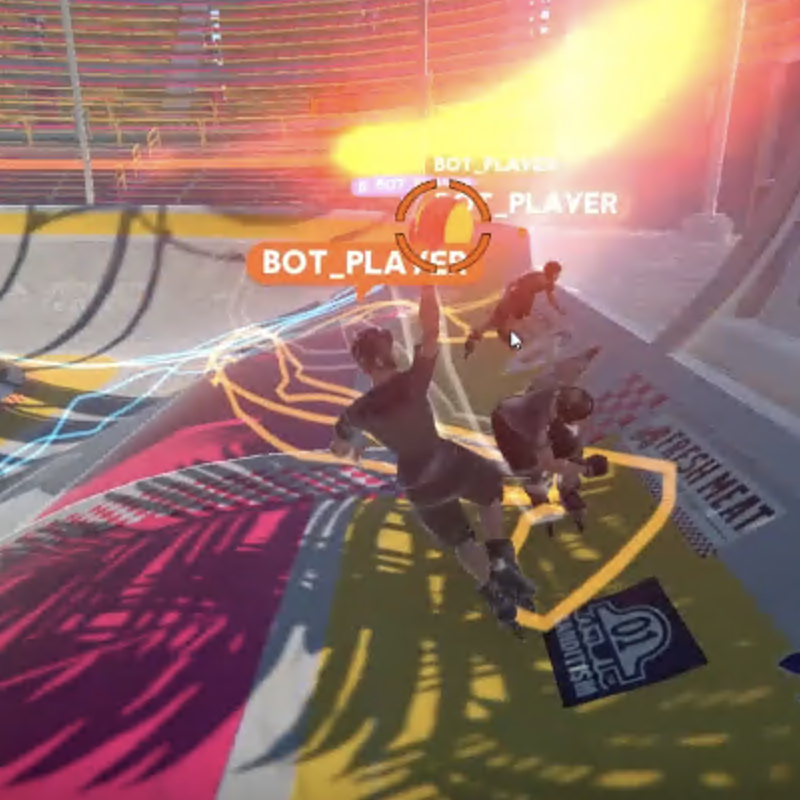}
         \caption{The Dodge action used to result in a sharp change in direction (as indicated by the orange trail). This was often used at checkpoint 1 (at the goal line) to switch from clockwise to anti-clockwise skating in order to progress checkpoints.}
         \label{fig:dodgeturn}
     \end{subfigure}
        \caption{The Draft mechanic and the Dodge action were exploited in earlier models before nerfing.}
        \label{fig:oldmechanics}
\end{figure}

In earlier models, all players of the same team tended to skate together throughout the match, as shown in figure~\ref{fig:3together}. We speculated this to be due to the \verb+Draft+ mechanic --whereby players skating in the wake of other players gain bonus speed-- being overpowered. This mechanic has since been nerfed. In recent models, the agents have a low tendency to skate together, opting for more strategic positioning instead (see section~\ref{sec:positions}).

An earlier version of the \verb+Dodge+ action caused a sharp change in direction (as shown in figure~\ref{fig:dodgeturn}), which was difficult to accomplish using normal navigation controls. The \verb+Dodge+ action was exploited by the agents to abruptly switch from clockwise to anti-clockwise skating and vice versa. Currently, \verb+Dodge+ causes the player to side-step while continuing in the same direction.

\subsection{Multi-purpose AIs}

The AIs are agnostic to game mode due to generic observations (section~\ref{sec:observations}). They are used in \verb+Classic+ game mode to replace human players who disconnect, and in \verb+Training with Bots+, which is a practice mode for one human player. \verb+3v3+ and \verb+2v2+ versions exist for both game modes. 

\begin{figure}[!h]
  \centering
  \includegraphics[width=0.6\linewidth]{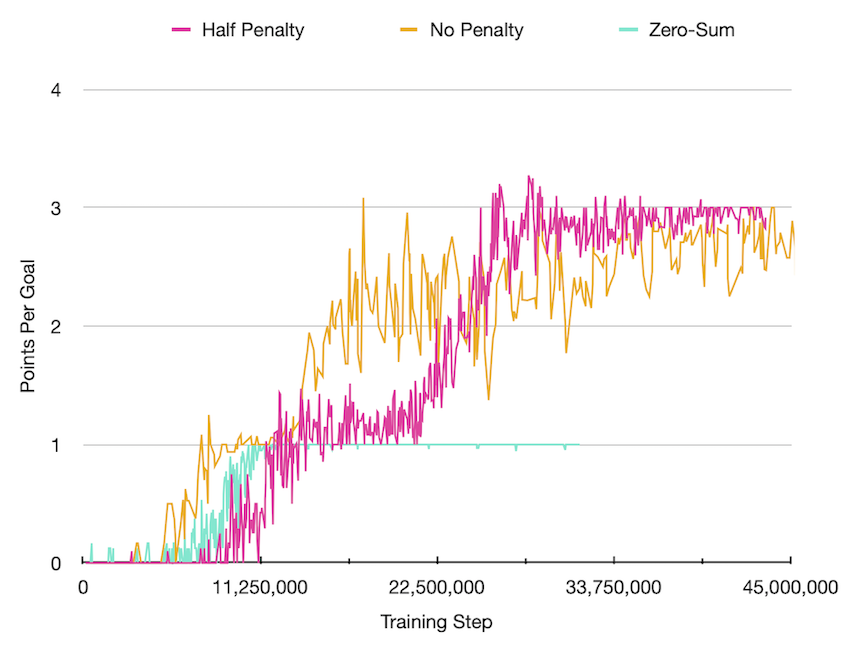}
  \caption{The penalty multiplier significantly affects the resulting behaviours. {\bfseries Zero-sum} rewards result in AIs that are cautious and defense-oriented-- it is clear from observing the model that multi-point goals are never attempted. {\bfseries Half Penalty} results in AIs that are balanced; frequently attempting multi-point goals while also developing sophisticated defense strategies. {\bfseries No Penalty} results in AIs that learn to score multi-point goals very quickly, but it takes them longer to develop good defense strategies.}
  \label{fig:points}
\end{figure}

Lower difficulty models are preferred for \verb+Classic+ game mode, while any difficulty level can be used in \verb+Training with Bots+. The penalty multiplier (section~\ref{sec:rewards}) can be used to obtain different difficulty levels. Figure~\ref{fig:points} shows how the penalty multiplier affects the number of points scored. 

\subsection{Emergent Behaviours}
\label{sec:emergentbhvrs}

The AIs consistently develop intelligent, co-ordinated, and co-dependent techniques and strategies. Some of the techniques developed by the agents were inline with the intent of the game designers, while others were exploits or simply off-brand uses of game mechanics. This section highlights some of the most commonly observed behaviours. Please refer to \href{https://vimeo.com/466780171}{https://vimeo.com/466780171}~\footnoteref{pass} for a video demonstrating some of these behaviours, along with longer goal sequences. 

\subsubsection*{Self-assigned Roles}
\label{sec:positions}

Agents are able to strategically position themselves in offensive or defensive positions depending on the game state.

\begin{figure}[!h]
     \centering
     \begin{subfigure}[b]{0.4\textwidth}
         \centering
         \includegraphics[width=\textwidth]{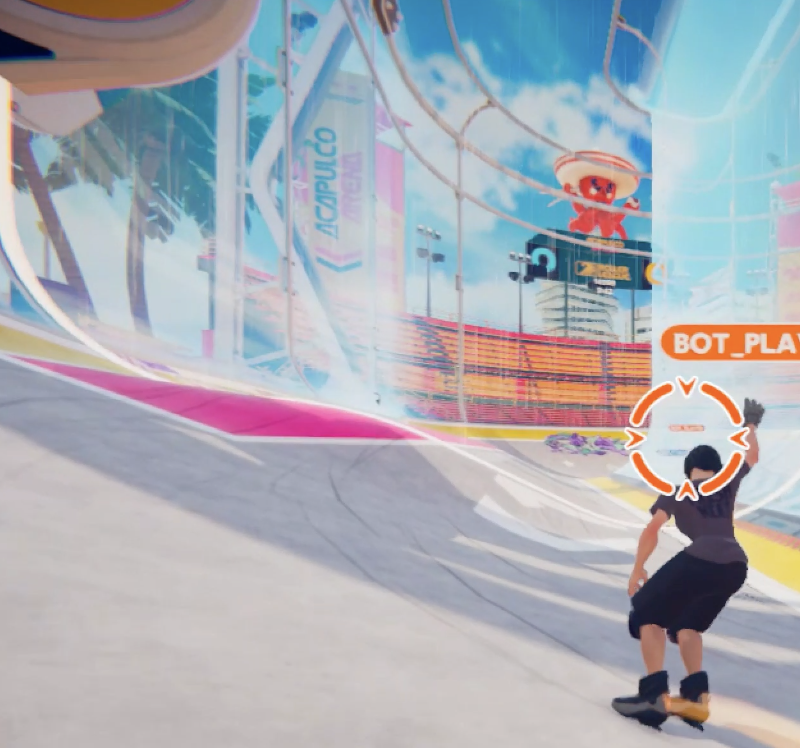}
         \caption{Orange agent is strategically positioned at the goal, performing Call for Pass as their teammate (indicated by orange ball marker) passes checkpoint 3 with the ball.}
         \label{fig:callforpassgoal}
     \end{subfigure}
     \hfill
     \begin{subfigure}[b]{0.4\textwidth}
         \centering
         \includegraphics[width=\textwidth]{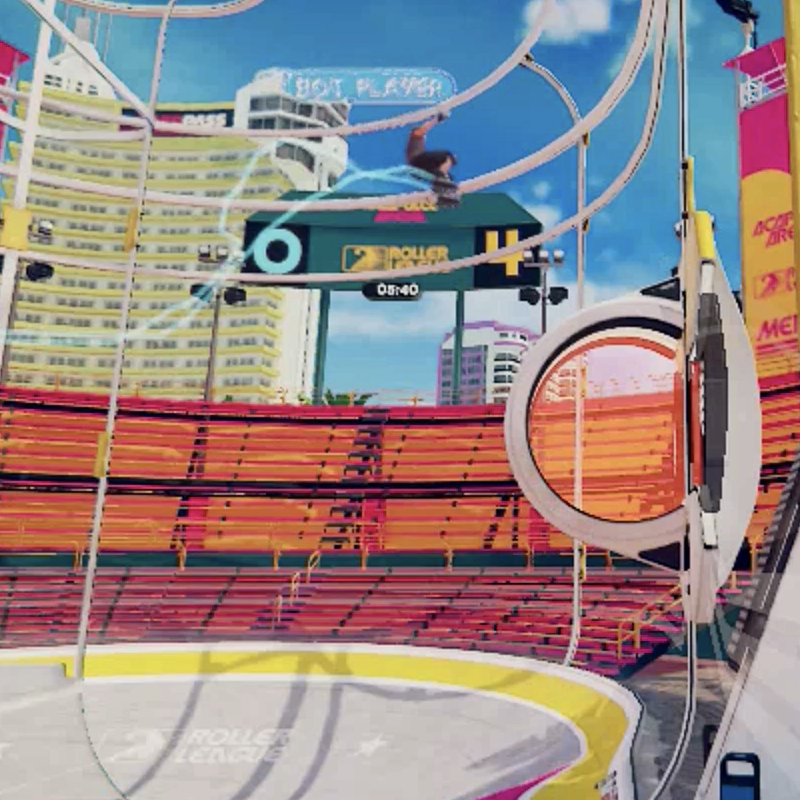}
         \caption{Blue agent goal tending by performing high jumps near the goal, while the goal is open for the Orange team.}
         \label{fig:interceptattempt}
     \end{subfigure}
        \caption{Strategic positioning depending on game state}
        \label{fig:positions}
\end{figure}

A commonly observed strategy is for \verb+Player A+ to progress \verb+checkpoints 1+ and \verb+2+, then pass the ball to \verb+Player B+ who is positioned at \verb+checkpoint 3+. \verb+Player A+ then goes back to the goal (at \verb+checkpoint 1+) to either \verb+Call For Pass+ and complete more laps, or to peel for \verb+Player B+ who proceeds to progress the final checkpoint before either passing to \verb+Player A+ or scoring. The defending players typically position themselves either at the goal or at \verb+checkpoint 3+ to intercept a pass or to tackle the receiving player. Figure~\ref{fig:positions} illustrates these positions.

\subsubsection*{Navigation}

\begin{figure}[!h]
     \centering
     \begin{subfigure}[b]{0.4\textwidth}
         \centering
         \includegraphics[width=\textwidth]{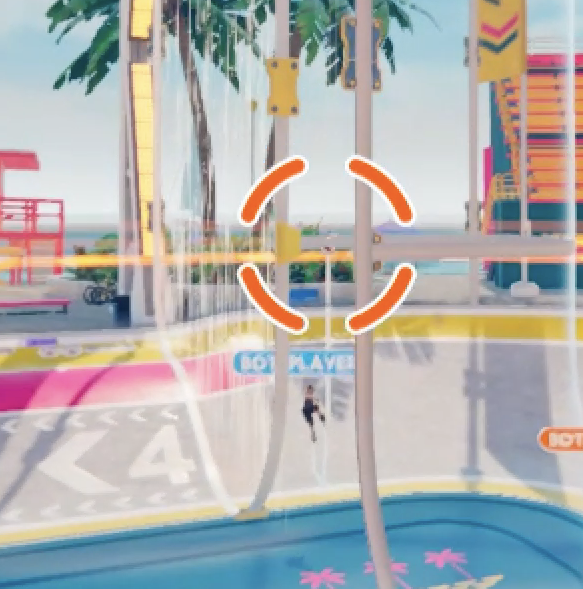}
         \caption{Blue agent using Jump to reach a ball in air (indicated by orange ball marker).}
     \end{subfigure}
     \hfill
     \begin{subfigure}[b]{0.4\textwidth}
         \centering
         \includegraphics[width=\textwidth]{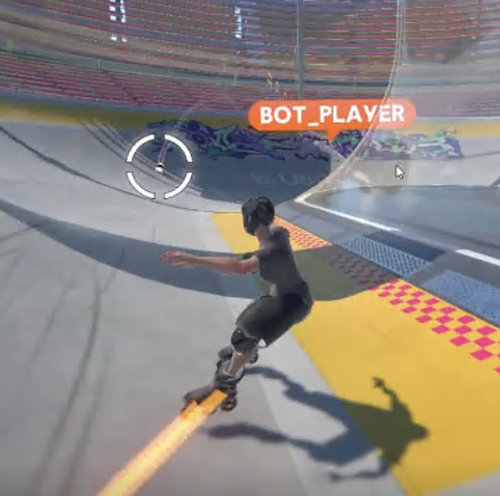}
         \caption{Using Brake to catch a ball rolling towards the agent}
     \end{subfigure}
        \caption{Agents can utilize multiple player actions to navigate to the ball depending on its relative position and speed}
        \label{fig:catchball}
\end{figure}

Agents use almost all available player actions and combos for navigation. Figure~\ref{fig:catchball} shows the agents using different player actions to reach the ball depending on its relative position and speed.

\begin{figure}[!h]
     \centering
     \begin{subfigure}[b]{0.4\textwidth}
         \centering
         \includegraphics[width=\textwidth]{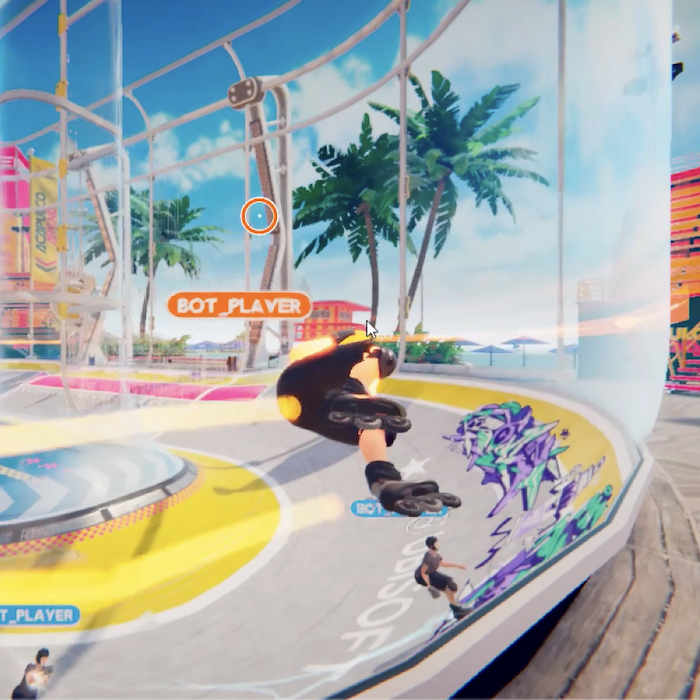}
         \caption{Orange agent skating on outer arena wall}
     \end{subfigure}
     \hfill
     \begin{subfigure}[b]{0.4\textwidth}
         \centering
         \includegraphics[width=\textwidth]{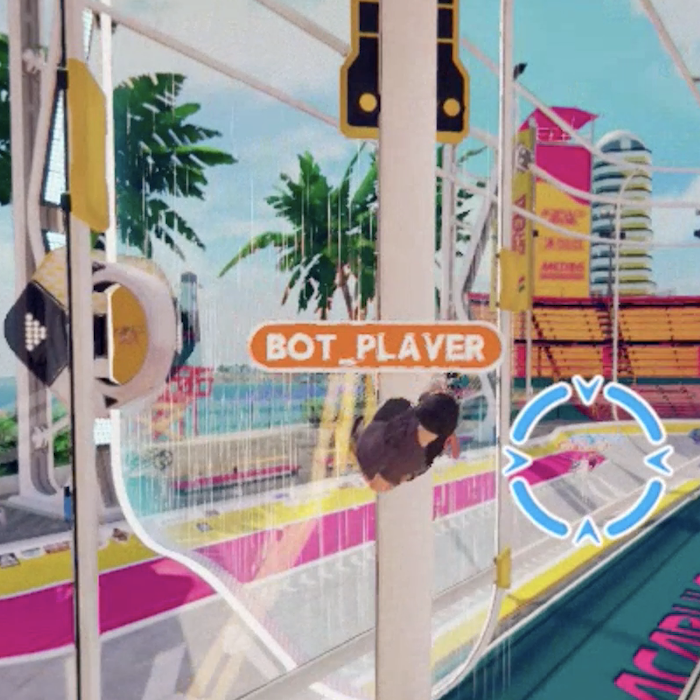}
         \caption{Orange agent skating on inner arena wall}
     \end{subfigure}
        \caption{Agents skate on the walls of the arena to avoid the enemy team}
        \label{fig:highwall}
\end{figure}

Figure~\ref{fig:highwall} shows agents skating on the walls of the arena to avoid the enemy team.
Earlier models exploited game mechanics for navigation, as discussed in section~\ref{sec:balance}.

\subsubsection*{Other}

\begin{figure}[!h]
     \centering
     \begin{subfigure}[b]{0.35\textwidth}
         \centering
         \includegraphics[width=\textwidth]{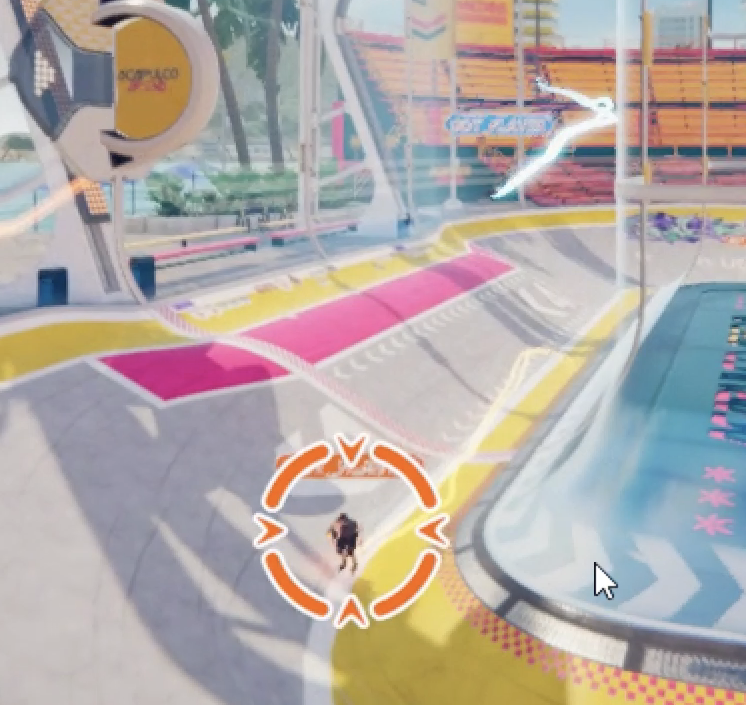}
         \caption{Blue agent performing a combo involving Uppercut and Dash while jumping off the wall to target an enemy}
     \end{subfigure}
     \hfill
     \begin{subfigure}[b]{0.45\textwidth}
         \centering
         \includegraphics[width=\textwidth]{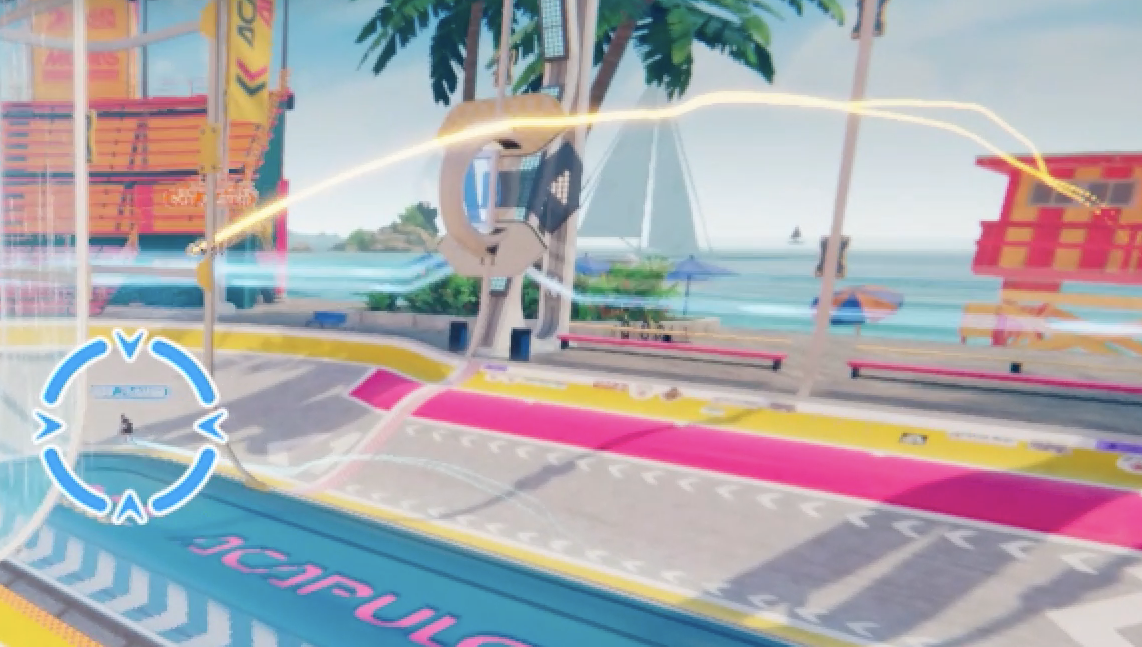}
         \caption{Orange agent performing a combo to target a far away enemy}
     \end{subfigure}
        \caption{Agents are able to effectively use combos to traverse the arena and tackle enemies}
        \label{fig:combos}
\end{figure}

Figure~\ref{fig:combos} demonstrates agents effectively using action combos to quickly traverse the arena and tackle enemies. Combos are made possible given the action space and observations described in sections~\ref{sec:observations} and~\ref{sec:actionspace}.

Even though the \verb+Pass+ mechanic automatically selects a target, agents sometimes favour the \verb+Throw+ action. For instance, when enemy players are in the way between the ball bearer and their target ally, the ball bearer passes the ball by bouncing it off the wall rather than passing it directly to their ally.

\FloatBarrier
\section*{Conclusion}

This work demonstrates that RL can be a powerful and practical AI design tool for competitive, cooperative, modern games. Our RL system for Roller Champions accomplished the following goals:

\begin{itemize}
\item The system is highly efficient memory and cpu-wise due to the use of a small policy, simple observations derived directly from the game state, and high-level directions in the action space. This makes it suitable for real-time inference.
\item The turnover for training is 1--4 days, making it suitable for a live game or a game that's still in development.
\item By keeping the rewards simple, the AIs readily adapt to changes in game mechanics, inadvertently identifying over-powered or under-powered mechanics.
\item The AIs consistently develop sophisticated co-operative strategies despite the system's relative simplicity and efficiency.
\end{itemize}

In conclusion, our system is capable of producing sophisticated game AIs with complex strategies and co-dependent co-operative behaviours, is practical for fast-paced agile development, and suitable for an online multiplayer game.



\FloatBarrier
\begin{ack}
The authors would like to thank Ubisoft and the Roller Champions team for enthusiastically embracing this project.
\end{ack}

\small
\bibliographystyle{IEEEtran}
\bibliography{rollerRL}
\clearpage

\end{document}